\documentclass[sigconf]{acmart}
\newtheorem{definition}{Definition}

\usepackage{listings}
\usepackage{xcolor}

\definecolor{codegreen}{rgb}{0,0.6,0}
\definecolor{codegray}{rgb}{0.5,0.5,0.5}
\definecolor{codepurple}{rgb}{0.58,0,0.82}
\definecolor{backcolour}{rgb}{0.95,0.95,0.92}

\lstdefinestyle{mystyle}{
    backgroundcolor=\color{backcolour},
    commentstyle=\color{codegreen},
    keywordstyle=\color{magenta},
    numberstyle=\tiny\color{codegray},
    stringstyle=\color{codepurple},
    breakatwhitespace=false,         
    breaklines=true,                 
    captionpos=b,                    
    keepspaces=true,                 
    numbers=left,                    
    numbersep=5pt,                  
    showspaces=false,                
    showstringspaces=false,
    showtabs=false,                  
    tabsize=2,
}
\usepackage{booktabs}						
\usepackage{multirow}						
\usepackage{amsfonts}						
\usepackage{graphicx}						
\usepackage{duckuments}

\AtBeginDocument{%
  }

\setcopyright{none}
\begin{document}

\title{GraphiT: Efficient Node Classification on Text-Attributed Graphs with Prompt Optimized LLMs}


\author{Shima Khoshraftar, Niaz Abedini and Amir Hajian}
\affiliation{%
  \institution{Arteria AI}
  \city{Toronto}
  \country{Canada}}

\renewcommand{\shortauthors}{Khoshraftar et al.}

\begin{abstract}
  The application of large language models (LLMs) to graph data has attracted a lot of attention recently.
LLMs allow us to use deep contextual embeddings from pretrained models in text-attributed graphs, where shallow embeddings are often used for the text attributes of nodes. However, it is still challenging to efficiently encode the graph structure and features into a sequential form for use by LLMs. In addition, the performance of an LLM alone, is highly dependent on the structure of the input prompt, which limits their effectiveness as a reliable approach and often requires iterative manual adjustments that could be slow, tedious and difficult to replicate programmatically.
In this paper, we propose GraphiT (\textbf{Graph}s \textbf{i}n \textbf{T}ext), a framework for encoding graphs into a textual format and optimizing LLM prompts for graph prediction tasks. Here we focus on node classification for text-attributed graphs. We encode the graph data for every node and its neighborhood into a concise text to enable LLMs to better utilize the information in the graph. We then further programmatically optimize the LLM prompts using the DSPy framework to automate this step and make it more efficient and reproducible. GraphiT outperforms our LLM-based baselines on three datasets and we show how the optimization step in GraphiT leads to measurably better results without manual prompt tweaking. We also demonstrated that our graph encoding approach is competitive to other graph encoding methods while being less expensive because it uses significantly less tokens for the same task.

\end{abstract}




\maketitle

\footnotetext{Corresponding author: shima.khoshraftar@arteria.ai}

\section{Introduction}
Graphs are powerful tools for representing entities and the relationships between them in different applications such as social networks and citation networks. For instance, in a citation network, nodes are the articles and there is an edge between two articles if one article cites another one. In text-attributed graphs, nodes have text attributes which provide further information about the nodes. In the citation network described above, the text attributes of a node could be the content of the associated article. One of the main applications of graphs is the node classification task in which a model predicts a label for the nodes in the test set.

Graph Neural Nets (GNNs) \cite{zhang2020deep, khoshraftar2024survey} are the state of the art in graph representation learning. They typically generate a node embedding by aggregating the embeddings of neighbors of the node in a message passing mechanism \cite{kipf2016semi, velivckovic2018graph}. GNNs consider the structure and the attributes of graphs in generating embeddings. The text attributes of nodes are often represented by shallow embeddings such as bag-of-words \cite{katz1985philosophy} and word2vec \cite{mikolov2013efficient} which can not capture the contextual relationships between words in text attributes. However, large language models have demonstrated great success in generating contextual text embeddings with superior performance than shallow embeddings in natural language processing (NLP) tasks. The success of the LLM models is mainly due to their pre-training on a vast amount of text corpora which gives them massive knowledge and semantic comprehension capabilities. Hence, many recent efforts have explored combining LLMs and GNNs \cite{he2023harnessing, duan2023frustratingly, yu2023empower, zhu2024efficient,xue2023efficient}.
While effective, this combination results in a complex system involving two large models which increase the computational demands and require labeled training data. 

Consequently, other studies focused on evaluating the potential of LLMs to act as standalone models for both embedding generation and prediction \cite{fatemitalk, ye2024language, perozzi2024let,chen2024exploring}.
 These methods employ various techniques for optimizing LLMs, which can be broadly categorized into prompt engineering \cite{fatemitalk} which relies heavily on manual adjustments or fine-tuning which require labeled training data \cite{ye2024language}. 
Additionally, different approaches are explored for converting graph structures into sequential formats suitable for LLMs, including using text attributes, lists of a node's neighbors \cite{ye2024language}, and neighbor summaries \cite{chen2024exploring} which can lead to increasing the context length of prompts making the LLM calls more expensive.


In this paper, we investigate the promise and limitations of using LLMs for the node classification tasks by proposing new approaches for graph encoding and prompt optimization in terms of instruction and examples using DSPy framework \cite{khattab2023dspy}. Specifically, we use a prompt programming approach which automates the optimization of LLMs for node classification without extra training, manual tweaks and with a small set of labeled data. 
Furthermore, 
we propose using keyphrases of neighbor nodes to represent a node, which offer several advantages. First, keyphrases require significantly less of the LLMs' context window while effectively conveying the key points. Second, when neighbor summaries are lengthy, LLMs may experience the "lost-in-the-middle" effect \cite{liu2024lost}, where critical information representing a node's neighbors is overlooked. Lastly, in certain graph applications, including multi-hop neighbors is essential. However, summarizing such extended neighborhood information becomes challenging and less interpretable. By using keyphrases, we can generate concise yet informative summaries that capture a broader span of information within the graph.
Our main contributions in this paper are as follows:
\begin{itemize}
    \item We present GraphiT, a novel technique for graph encoding and LLM prompt optimization in node classification task.
    \item GraphiT provides an efficient solution for minimizing the use of LLM context window and automating LLM prompt optimization.
    \item We evaluate the performance of our approach with three baselines on three datasets. In addition, we perform ablation studies to show the effectiveness of GraphiT components. GraphiT can be easily adapted to new tasks and datasets with minimal effort. 

\end{itemize}

\section{Related works}
GNNs are the frontier techniques in the field of graph representation learning \cite{zhang2020deep, khoshraftar2024survey}. However, they use shallow embeddings to represent text attributes of nodes. Given the capability of LLMs to generate rich contextual embeddings, several works have combined LLMs with GNNs to enhance GNN performance \cite{he2023harnessing, duan2023frustratingly, yu2023empower, zhu2024efficient,xue2023efficient}. Leading \cite{xue2023efficient} employs an end-to-end training of LMs and GNNs for graph prediction tasks. Engine \cite{zhu2024efficient} combines LLMs and GNNs using a tunable side structure. Despite their effectiveness, these integrations create complex systems that are often computationally intensive and require labeled data for training.
Therefore, other studies investigate the possibility of using LLMs alone for graph prediction tasks. 
In \cite{wang2024can}, LLMs are utilized for several graph reasoning tasks such as connectivity, shortest path and topological sort using two instruction-based prompt engineering techniques. InstructGLM \cite{ye2024language} proposes a instruction fine-tuning method for node classification by LLMs. In \cite{chen2024exploring}, LLMs have been used both as enhancer and predictor for node classification task. It encodes the nodes into text by incorporating text attributes and 2-hop neighbors summaries. In \cite{fatemitalk}, different methods for graph encoding and prompt engineering were investigated. In \cite{perozzi2024let}, graphs are input to LLMs using a graph encoder which was trained similar to soft prompting methods \cite{lester2021power}. 
Fine-tuning and soft promoting techniques require training with labeled data. Traditional prompt engineering relies heavily on human expertise and manual adjustments. In contrast, we programmatically optimize LLM usage with only a small set of labeled data. In addition, we efficiently capture the information in a node's neighborhood by extracting keyphrases from text attributes of neighboring nodes. 



\section{Method}
\subsection{Problem definition}
Let $G=(V, E, S)$ be a text-attributed graph $G$ where $V$, $E$ and $S$ represent nodes, edges and text attributes of nodes in the graph, respectively. For each node $v_i \in V$, $s_i \in S$ represents the text attributes of $v_i$. $Y$ is the set of labels associated with nodes. Our goal is to perform node classification on the graph using a large language model. In the node classification, a label is predicted for each node in the graph. Formally, a classifier $f$ maps the set of nodes $V$ to the set of labels $Y$ represented as:  $f: V \rightarrow Y$. 
The core of our approach consists of three main steps: 1) each node $v_i$ in the graph is encoded into a sequential form for use by LLM, 2) an LLM prompt is optimized in terms of instruction and demonstrative examples. 3) the LLM with the optimized prompt is utilized to assign a label to each node.

\begin{figure}
    \centering
    \includegraphics[scale=0.6]{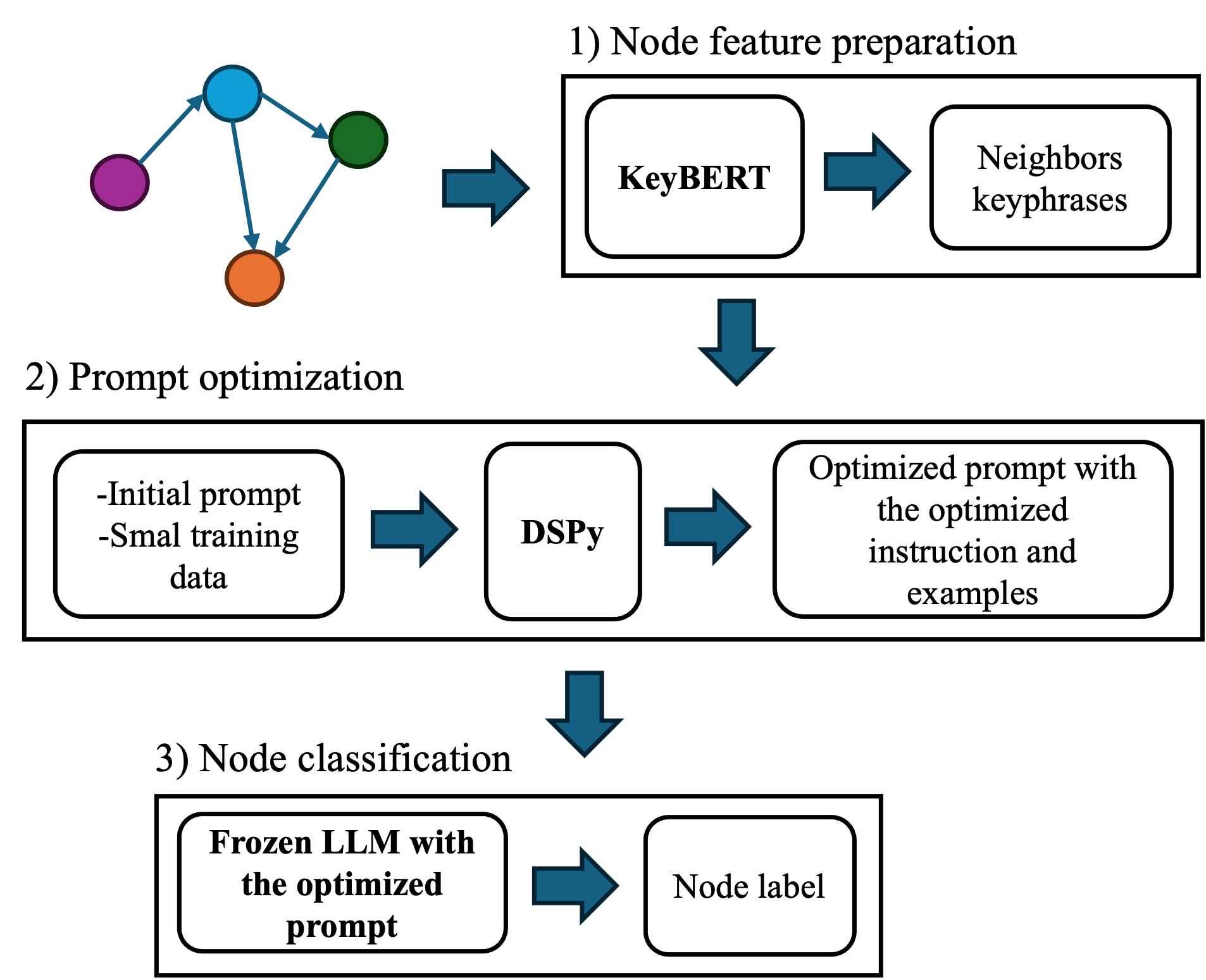}
    \caption{The general framework of GraphiT. First, node features, including neighbors keyphrases, are extracted for each node in the graph. Next, a small subset of nodes, along with an initial prompt, are fed into DSPy to produce an optimized prompt. Finally, node classification is performed using the optimized prompt.}
    \label{framework}
\end{figure}


\subsection{Node feature preparation}
\label{sec:Node feature preparation}
While LLMs have shown remarkable success with textual data, a crucial question remains: how LLMs can best utilize the information in structured graph data?
\cite{perozzi2024let}. In this study, we use the homophily assumption in graphs which says connected nodes are similar \cite{ciotti2016homophily} and for each node consider the features of the 1-hop neighbors of a node to help the LLM in predicting the node labels. For a node $v_i$ with $N_i = \{v_0,...,v_k\}$ representing the set of its 1-hop neighbors, $v_0$ to $v_k$, we consider the following features beside node text attributes.


\begin{definition}
(Neighbors labels). This set consists of labels of 1-hop neighbors of node $v_i$ denoted as $l_{N_i} = \{l_0,...,l_k\}$. 
\end{definition}


\begin{definition}
(Neighbors keyphrases). Let $s_{N_i} = \{s_0,...,s_k\}$ be a set consisting of text attributes of 1-hop neighbors of the node. 
Neighbors keyphrases denoted by $p_{N_i}= \{p_0,...,p_\zeta\}$ is a set containing the $\zeta$  keyphrases that are shared among the node's neighbors text attributes.
\end{definition}

The process for extracting neighbors keyphrases for each node is detailed in the next section. We apply the keyphrase extraction algorithm to the concatenation of elements in $s_{N_i}$.

\subsubsection{Keyphrase extraction}
\label{sec:keyphrase} 
Keyphrase extraction (KPE) is an automated process that identifies the most important words or phrases from a given text. These keyphrases 
are useful for various downstream applications, such as document classification, clustering, summarization, indexing documents, query expansion, and interactive document retrieval.
Various approaches have been developed for efficient extraction of keyphrases over the past three decades. See \cite{Papagiannopoulou2019ARO} and the references therein for a comprehensive review of well-established keyphrase extraction methods and \cite{Schopf2022PatternRankLP} for a more recent approach using semantic extractions. In our current work, we use a semantic KPE approach that works as follows: first $n$-grams are generated using a count vectorizer using the optimal choice of $n$ identified experimentally. The candidate keyphrases are then mapped into dense representations using Transformer-based contextual embeddings. Similar keyphrases are identified through an embedding-based semantic similarity measure and ranked based on their similarity scores to the input text {\cite{DBLP:journals/corr/abs-1801-04470} using KeyBERT implementation by \cite{grootendorst2020keybert}. The top-ranked candidates are considered the most relevant keywords/keyphrases for the text. Then a diversity module is applied to the selected keyphrases to ensure redundancy reduction using techniques like maximum marginal relevance and max sum similarity \cite{DBLP:journals/corr/abs-1801-04470}. The final result is a set of most important keyphrases that encompass the main content of the input text.


An example of node neighbors summary and node neighbors keyphrases is shown in Table \ref{example-table}. In this example, we see that the keyphrases capture the main concepts of the neighbors summary in a concise manner.

\begin{table}[]
\caption{Example of extracted node neighbors keyphrases compared to the node neighbors summary for a random node with one neighbor in the Cora dataset. }
\label{example-table}
\begin{tabular}{lp{4.5cm}}
\hline
\textbf{Neighbors summary}    & This paper presents an algorithm using reinforcement learning at each node for packet routing in networks; it utilizes local information and outperforms traditional methods with minimal routing times through experiments, even in irregularly-connected network structures. \\ \hline
\textbf{Neighbors keyphrases} & \begin{tabular}[c]{@{}l@{}}distributed reinforcement learning, \\
network routing,\\
routing policies,\\
packet routing \end{tabular} \\ \hline
\end{tabular}
\end{table}


\subsection{Prompt optimization}
Considering that the quality of the input prompt to an LLM has a huge effect on the output of the model, we optimize the LLM prompt for the node classification both in terms of instruction and examples. In order to do that, we use the optimization framework of DSPy programming model. DSPy provides a framework in which we can define our task as a program and automatically optimize the prompt for the best performance. 
We will explain each step in the following sections.

\subsubsection{Node classification program} 
The program for node classification is illustrated in Code Snippet \ref{program}. Given the node features and a set of node labels as options, an LLM predicts a label for each node. We use the chain of thought technique \cite{wei2022chain} to let the LLM solve the problem step by step by breaking down the question into simpler tasks. 

\renewcommand{\lstlistingname}{Code Snippet} %
\begin{lstlisting}[caption={\texttt{DSPy} code for NodeClassification program with minor alterations for brevity.}, label={program}, language=Python,breaklines=true,showstringspaces=false,literate={í}{{\'i}}1]
class NodeClassification(dspy.Module):
  def __init__(self):
    self.cot = dspy.ChainOfThought(NodeClassficationSignature)
  
  def forward(self, NodeFeatures: str, options: list[str]) -> Prediction:
    # Predict with LM
    output = self.cot(NodeFeatures=NodeFeatures, options=options).completions.output
    return dspy.Prediction(predictions=output)
\end{lstlisting}

\subsubsection{Signature} 
The prompt for the LLM in the node classification program is defined using a signature abstraction. The signature for node classification on Cora and PubMed datasets are defined in Code Snippet \ref{signature}. This signature contains a task description in a docstring, the node features and a set of labels as inputs and the predicted output along with description and formatting information. As we encode the node information into a text format for use by LLM, we also formulate the node classification task into a text classification task in the task description and ask the LLM to classify a given text into the most applicable category. 
As Cora and PubMed are citation datasets, the task description specifies that the text is a scientific paper but this can be adjusted for any new dataset depending on the dataset graph content.

\renewcommand{\lstlistingname}{Code Snippet} %
\begin{lstlisting}[caption={\texttt{DSPy} signature of NodeClassification program for Cora and PubMed.}, label={signature}, language=Python,breaklines=true,showstringspaces=false,literate={í}{{\'i}}1]
class NodeClassficationSignature(dspy.Signature):
    __doc__ = f"""Given a snippet from a scientific paper, pick the most applicable category from the options."""
 
    NodeFeatures = dspy.InputField(prefix="Paper:")
    options = dspy.InputField(
        prefix="Options:",
        desc="List of comma-separated options to choose from",
        format=lambda x: ", ".join(x) if isinstance(x, list) else x,
    )
    output = dspy.OutputField(
        prefix="Category:"
    )

\end{lstlisting}

\subsubsection{Compilation}
We optimize the node classification program in terms of instruction and prediction examples. DSPy compilers handle this optimization programmatically. For instruction optimization, we use COPRO (Coordinate-ascent Optimization by Prompting) \cite{khattab2023dspy, opsahl2024optimizing}, an extension of OPRO approach \cite{opro}. 
The OPRO method relies on LLMs to iteratively optimize their own prompt based on a given problem description. COPRO generalizes OPRO \cite{opro} by incorporating a coordinate ascent strategy, allowing it to be applied to programs with multiple prompts. In this approach, each prompt is optimized individually while the other parameters remain fixed. In DSPy, the compiler continuously refines the program's instructions based on the LLM’s performance on the validation set, ultimately converging to a set of optimized instructions tailored to the task. 

Similarly, a set of optimized demonstrative examples are added to the prompt by an iterative process using bootstrap few-shot random search approach \cite{khattab2023dspy, opsahl2024optimizing}. In this process, 
a prediction is generated for each example within the training set. Let $\phi(x)$ represent the prediction for an example $x$, $x'$ denote the ground truth, and $\mu(\phi(x), x')$ be the score of the prediction compared to the ground truth based on a metric  $\mu$. If $\mu(\phi(x), x') \ge \lambda$ where $\lambda$ is a predefined threshold, the prediction is considered successful. Upon successful prediction, a demonstration comprising the input to the LLM and the corresponding output is recorded. A predetermined maximum number of these demonstrations are then incorporated into the prompt.
This process is repeated multiple times and the most performant demonstrations on the validation set are selected through random search.
We measure the performance of each program using the rank-precision at top K results (RP@K) and the metric defined as \cite{d2024context}:
\begin{equation}
    RP@K = \frac{1}{N} \sum_{n=1}^{N}\frac{1}{min(K,R_n)}\sum_{k=1}^{K}Rel(n,k)
\end{equation}
where $R_n$ is the set of labels for a node $n$, $Rel(n,k)$ is $1$ if the $k$-th predicted label for node $n$ is relevant and otherwise is $0$. $N$ is the total number of nodes in the set. 
\begin{table}[]
\centering
\caption{The node classification results in terms of accuracy}
\label{results2}
\begin{tabular}{l|llll}
\cline{1-4}
Method   & Cora  & PubMed & Ogbn-arxiv &  \\ \cline{1-4}
Vanilla    & 74.49 & 87.56  & 49.5 &  \\
Chen et al \cite{chen2024exploring} & 74    & 90.75  & 55    &  \\

GraphiT  & 79.84  & 93.28  & 57.25  &  \\ \cline{1-4}
GCN \cite{kipf2016semi}  & 82.20 & 81.01  & 73.10 &  \\ \cline{1-4}
\end{tabular}
\end{table}

\begin{table*}[]
\centering
\caption{Node classification results from our experiments with different node encoding techniques for GraphiT and a Vanilla LLM.} 
\label{results}
\begin{tabular}{l|l|ccc}
\hline
Dataset                 & Graph info                                  & Vanilla & GraphiT  \\ \hline
\multirow{4}{*}{Cora}   & Text attributes                             & 57.65  & 59.18 \\
                        & Text attributes + Neighbors labels               & 71.68  & 78.31 \\
                        & Text attributes + Neighbors labels + Neighbors summary & 72.19  & 80.1\\
                        & Text attributes + Neighbors labels + Neighbors keyphrases & 74.49  & 79.84  \\ 
                        
                        \hline
\multirow{4}{*}{PubMed} & Text attributes                             & 89.55  & 93.03\\
                        & Text attributes + Neighbors labels               & 87.06  & 90.54  \\
                        & Text attributes + Neighbors labels + Neighbors summary & 87.31  & 92.78 \\
                        & Text attributes + Neighbors labels + Neighbors keyphrases & 87.56  & 93.28 \\
                         
                         \hline
\multirow{4}{*}{Ogbn-arxiv} & Text attributes                             & 40  &  45.25\\
                        & Text attributes + Neighbors labels               &  49.75 &  55\\
                        & Text attributes + Neighbors labels + Neighbors summary & 49 & 58.5 \\
                        & Text attributes + Neighbors labels + Neighbors keyphrases &49.5&  57.25\\
                         
                         \hline

\end{tabular}
\end{table*}
\section{Experiments}
\subsection{Datasets}
    
We evaluated GraphiT was on three public datasets: Cora, PubMed and Ogbn-arxiv. Cora ~\cite{he2023harnessing} is a citation network where each node is an article and each edge indicates a citation relationship between two articles. Number of nodes and edges are 2708 and 5429. Each node belongs to one of the 7 classes: case based, genetic algorithms, neural networks, probabilistic methods, reinforcement learning, rule learning, and theory. Each node is associated with a text attribute containing the title and the abstract of the article. Similarly, PubMed ~\cite{he2023harnessing} is a citation network with 19k nodes and 44k edges. Each node in the dataset has one of the three labels: experimental induced diabetes, type 1 diabetes, and type 2 diabetes. The text attributes of nodes in PubMed are similar to Cora. Ogbn-arxiv \cite{hu2020open} is also a citation networks between all Computer Science arxiv papers containing 169k nodes, 1M edges and 40 subject areas.

\subsection{Settings}
Similar to ~\cite{chen2024exploring}, we randomly selected 200 nodes from the test set of each dataset as our test data. 
The reported scores are averaged over two sampled test sets. Our evaluation metric is $RP@1$ which is equivalent to accuracy in our experiments. The LLM that we used was \lstinline|gpt-3.5-turbo-1106|. We used \lstinline|BootstrapFewShotWithRandomSearch| and  \lstinline|COPRO| compilers from DSPy. The length of ngrams in the keyphrase extraction step is set to $ngram \in \{1,2,3\}$ and we set $\zeta=5$. The nodes neighbors summaries are generated using the quantized version of the Phi 3.5 model \cite{abdin2024phi3technicalreporthighly, phi3-quantized} by llama.cpp \cite{llama.cpp}.

\subsection{Node classification}
We evaluate the performance of GraphiT compared to three baselines on three datasets. Each node in the graph is encoded by integrating the node's neighbors' keyphrases with its text attributes and the labels of its neighbors. Without the loss of generality, for prompt optimization, we generate small training and validation sets by randomly sampling 3 and 2 nodes per class from training and validation sets of Cora dataset. Then, we use the optimized programs for inference on arbitrarily large test sets. 

Table \ref{results2} presents the node classification results from GraphiT, the result from the vanilla LLM using the same graph encoding, the best results from an unoptimized few-shot learning approach using LLMs by Chen et al \cite{chen2024exploring} and a graph convolutional network (GCN) result \cite{kipf2016semi} obtained from \cite{chen2024exploring}. We were able to compare with the methods reported in \cite{chen2024exploring} as they used the same number of nodes in the test sets for each dataset as us and were designed for the node classification task. GraphiT outperforms the results by the LLM-based models on all three datasets. It also achieves superior performance on PubMed compared to GCN. However, GraphiT falls short of the performance by GCN on Cora and Ogbn-arxiv datasets. This could be because GCN captures information from 2-hop neighbors for each node, which is useful for node classification on those datasets. Exploring the incorporation of neighbors beyond 1-hop in GraphiT will be one of our future research directions.


\subsection{Ablation study}
We investigate the effects of different components of GraphiT across three dataset. One major component of our model is the node neighbors keyphrases. We consider four settings to encode nodes into a sequence format, beginning with only the text attributes of the nodes and progressively incorporating additional features through concatenation. Table \ref{results} shows the results of the node classification for GraphiT in four settings. For all the datasets, incorporating neighbors keyphrases alongside the text attributes and neighbors labels enhances performance. Moreover, this approach has a comparable or better results compared to using neighbor summaries while significantly reducing the context length in the LLM prompt.  In Table \ref{results}, we also have the node classification results from Vanilla LLM across the four settings. Using neighbors keyphrases has a similar effects on Vanilla LLM. Additionally, we can see that GraphiT consistently outperforms the Vanilla method across all node encoding techniques.

\subsection{Cost comparison of using neighbors keyphrases versus summary}
  In Figure \ref{hist}, we present a histogram depicting the ratio of the number of tokens in neighbors summary to those in neighbors keyphrases for all datasets combined. The figure indicates that the average number of tokens resulting from the KPE approach on the node neighbors text is a few times smaller that the ones from the summarization method. 
 As a result, leveraging keyphrases leads to lower LLM API costs while still delivering competitive results compared to the summarization approach, as shown in Table \ref{results}.
In addition, for the KPE method, we use small encoder models for the generation of embeddings which is fast and lightweight, easily suitable for running on ordinary CPUs of today's laptops \cite{reimers-2019-sentence-bert}.

\begin{figure}
    \centering
    \includegraphics[scale=0.6]{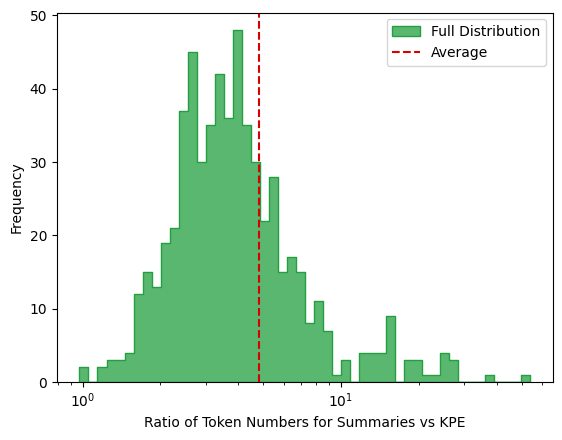}
    \caption{Histogram of the ratio of the number of tokens for summaries to those obtained from the KPE approach. The KPE method applied to the node neighbors results in significantly less tokens compared to the summarization method with minimal impact on the quality of the classification results. This reduction translates to lower LLM API costs by making the input context length considerably shorter.  }
    \label{hist}
\end{figure}

\section{Conclusions}
Our paper focuses on graph encoding and LLM optimization for the node classification task on text-attributed graphs. We demonstrate that the information in the nodes neighborhood is efficiently represented by the right choice of keyphrases. In addition, we optimize the LLM prompt automatically by refining instructions and adding demonstrative examples to the prompt leveraging the DSPy optimization framework. 
We compare the performance of our approach, GraphiT, with three baselines across three public datasets. The results demonstrate that our approach has a better performance compared to other models that are based on LLM models in all experiments. While promising for optimizing LLMs in the node classification task, GraphiT falls short of the GNNs performance on two datasets, highlighting a key area for our future research. Strategies like incorporating more neighborhood information for a node and integrating LLMs with GNNs could help bridge this performance gap. Furthermore, we will extend our approach to other graph prediction tasks, including link prediction.

\bibliographystyle{ACM-Reference-Format}
\bibliography{sample-base}


\end{document}